\documentclass[twoside,11pt]{article}

\usepackage[abbrvbib, preprint]{jmlr2e}
\usepackage{amsmath}
\usepackage{mathtools}

\usepackage{microtype}
\usepackage{subfigure}
\usepackage{booktabs} 
\usepackage{multirow} 

\firstpageno{1}

\begin{document}

\title{Functional Network: A Novel Framework for Interpretability of Deep Neural Networks}

\author{\name Ben Zhang \email zhangben@zju.edu.cn\\
       \addr State Key Lab. of CAD \& CG\\
       School of Mathematical Sciences\\
       Zhejiang University\\
       Hangzhou, Zhejiang Provence, China
       \AND
       \name Zhetong Dong  \email ztdong@zju.edu.cn\\
       \addr State Key Lab. of CAD \& CG\\
       School of Mathematical Sciences\\
       Zhejiang University\\
       Hangzhou, Zhejiang Provence, China
	   \AND
	   \name Junsong Zhang  \email zhangjs@xmu.edu.cn\\
	   \addr Fujian Key Lab. of Brain-Inspired Computing Technique and
	   Applications\\
	   Department of Cognitive Science\\
	   Xiamen University\\
	   Xiamen, Fujian Province, China
	   \AND
	   \name{Hongwei Lin \thanks{Corresponding author}}\email hwlin@zju.edu.cn\\
	   \addr State Key Lab. of CAD \& CG\\
	   School of Mathematical Sciences\\
	   Zhejiang University\\
	   Hangzhou, Zhejiang Provence, China}

\editor{Kevin Murphy and Bernhard Sch{\"o}lkopf}

\maketitle

\begin{abstract}
The layered structure of deep neural networks hinders the use of numerous 
analysis tools and thus the development of its interpretability.
Inspired by the success of functional brain networks, 
we propose a novel framework for interpretability of deep neural networks,
that is, the functional network.
We construct the functional network of fully connected networks
and explore its small-worldness.
In our experiments, the mechanisms of regularization methods, 
namely, batch normalization and dropout, are revealed 
using graph theoretical analysis and topological data analysis.
Our empirical analysis shows the following:
(1) Batch normalization enhances model performance by 
increasing the global efficiency and the number of loops
but reduces adversarial robustness by lowering the fault tolerance.
(2) Dropout improves generalization and robustness of models 
by improving the functional specialization and fault tolerance.
(3) The models with different regularizations can be clustered correctly 
according to their functional topological differences,
reflecting the great potential of the functional network and 
topological data analysis in interpretability.
\end{abstract}


\section{Introduction}
\label{intro}
Deep neural networks are considered black-box models without a
sufficient level of interpretability,
which limits their wider applications and further development.
Some studies focused on the structural information to explain them.
However, many limitations exist:
(1) Using only the structural information is insufficient to explain
the performance differences of a model on diverse datasets.
(2) The connections between neurons are preset and fixed,
hindering the use of rich tools in network science.
(3) The layered network structure only depicts the interactions 
between the neurons in the adjacent layers, 
rather than those in the same and non-adjacent layers.
Hence, a new interpretable method with a general form 
that focuses on the network functions is urgently required.

Introducing the methods on brain function explanation 
in deep learning is feasible
because many similarities exist between deep neural networks and the brain:
(1) Deep learning is an artificial neural network technology inspired 
by the brain.
(2) Reports show that similar coding mechanisms exist between them
\citep{Yang2019Task,Bi2020Understanding}.
(3) Deep neural networks have been used as the computational models 
of the primate brain to explain its information processing
\citep{Yamins2014Performance,Cadieu2014Deep,Umut2015Deep}.
An important method to understand the brain is to build a functional
brain network that describes the statistical dependencies among
neural activities of brain regions
\citep{Carolyn2018Functional,Beaty2018Robust}.

Motivated by the functional brain network,
we propose a novel framework for interpretability of deep neural networks, 
that is, the functional network.
This network can maintain practicability and provide insights into neuroscience.
Given a deep neural network and a dataset, 
we record the output values of neurons when the model processes the dataset, 
compute the statistical dependencies among them,
and construct the functional network by network binarization.
In contrast to the structural network, the functional network 
depicts the functional interactions among neurons 
in the same layer and non-adjacent layers, 
in addition to the adjacent layers.
By constructing the functional network, 
we introduce the powerful \emph{graph theoretical analysis} (GTA) and 
\emph{topological data analysis} (TDA) in the complex brain network analysis 
into interpretability of deep neural networks 
to capture the topological properties and high-order structures
to explain how the models work.	
Furthermore, when a neural network processes diverse datasets, 
various functional networks are generated, 
and the variations in the functional networks can 
be utilized to explain the performance differences.

In this work, we partly reveal the mechanisms of the 
\emph{fully connected network} (FCN).
The results show that,
similar to the functional brain network,
the functional network of FCNs is a small-world network.     
This result suggests that deep neural networks have a similar
functional organization to the brain,
in which information transmits efficiently at a low cost.
As an application of the functional network,
we quantitatively analyze the effects of commonly used regularization
techniques, namely, batch normalization and dropout, using graph theoretical and 
topological methods
and explain how the methods work.
Moreover, according to the topological differences between functional networks,
the models with different regularizations
can be correctly clustered.
These findings demonstrate that the functional network can not only provide
explanations for deep neural networks
but also evaluate the models in practice.

\section{Related Works}
\label{relatedwork}

In this section, we introduce the previous works on
interpretability of deep neural networks,
particularly those using TDA,
and the functional brain network in neuroscience.

\textbf{Interpretability of Deep Neural Networks}
In recent years, interpretability of deep learning has attracted 
increasing attention from researchers.
Several methods have been reported to address this issue,
such as the extraction of logical rules or decision trees
\citep{Boz2002Extracting,Richi2009Generating},
the interpretation for the semantics of neurons or
convolutional layers
\citep{Bau2017Network,Dalvi2019What,Zeiler2014Visualizing},
the local perturbation-based explanations
\citep{STRUMBELJ2009886,2016why,DBLP:conf/aaai/AkulaWZ20},
the prototype selection 
\citep{10.1214/11-AOAS495,10.5555/2969033.2969045},
and the generalization capability or complexity measures
\citep{Rieck2019Neural,Zhang2021Understanding,Maithra2017On}.
However, some disadvantages exist in the previous works.
For instance, the prototype selection and semantic interpretation
only focus on the single data or feature and
cannot provide a global understanding.
Moreover, the rule extraction is only appropriate for 
deep neural networks with few neurons.
Functional networks can overcome these shortcomings and explain 
the network mechanisms globally from the perspective of 
the functions of neurons.

\textbf{TDA and Interpretability}
Recently, several attempts have been made to apply TDA to study
interpretability of deep learning.
Naitza et al. investigated the changes of data topology in the working
process of FCNs
and found that FCNs work by simplifying data topology
until it becomes linearly separable \citep{Naitzat2020Topology}.
Rieck et al. proposed neural persistence to estimate 
the structure complexity and generalization ability 
using the zero-dimensional topological features 
\citep{Rieck2019Neural}.
Watanabe et al. extracted the one-dimensional topological structure
features to investigate the inner representations of FCNs
\citep{Watanabe2021Topological}.
These studies indicated that TDA can extract high-order topological
information to explain neural networks.
However, some methods only use the structural information of
the deep neural network and are not combined with data and tasks
\citep{Rieck2019Neural,Watanabe2021Topological}.
Some approaches are only applicable to the model with a small number of
neurons \citep{Naitzat2020Topology}.
In our work, TDA for the functional network 
depicts the functional organization of a deep neural network
and evaluates its topological properties globally.

\textbf{Functional Brain Network Analysis}
Using non-invasive brain-observation technologies,
such as fMRI, 
researchers can record the neural activities of brain regions,
calculate the statistical dependencies among them as the functional
connectivities,
and model the brain as a sparse binary graph,
called the functional brain network.
The nodes represent brain regions,
and the edges represent functional connectivities.
GTA is a powerful mathematical tool in complex brain network analysis.
GTA is used to describe and interpret brain changes during development
\citep{Menon2013Developmental},
reveal learning mechanisms
\citep{Bassett2011Dynamic,Bassett2015Learning},
understand the pathogenesis of brain diseases
and provide imaging biomarkers for diagnosis
\citep{Rudie2013Altered,Carolyn2018Functional}.
Nevertheless, the graph can only model the binary relation
and cannot model the multivariate relation in the brain.
Selecting a favorable threshold for functional network construction
is difficult.
To address these problems,
TDA, a rapidly developing mathematical tool based on algebraic topology,
is employed to assess the brain structures
\citep{Singh2008Topological,Petri2014Homological},
distinguish the brain states \citep{Billings2021Simplicial},
discover spatial coding principles \citep{Dabaghian2012A},
and study the pathogenesis of diseases \citep{Shnier2019Persistent}.
Those studies show that TDA can effectively describe higher-order interactions
and capture more topological information about functional organizations
in the brain without selecting a threshold.
In our work, we used GTA and TDA to reveal the mechanisms of
deep neural networks.

\section{Background: Graph Theoretical Analysis and Topological Data Analysis}
\label{background}
\subsection{GTA}
\label{APP:GTA}
Graph theory is the main mathematical tool in the field of 
complex network analysis.
A complex network is modeled as a binary graph model $G(V,E)$,
where $V=\{v_{i}\}_{1\leq i \leq n}$ is the node set and
$E=\{e_{k}|e_{k}=(v_{i},v_{j}),v_{i},v_{j}\in V\}_{1\leq k \leq m}$ is the edge set.
Edges can be weighted, and the weight function
$\psi: E\rightarrow \mathbb{R}^{+}\cup \{0\}$
maps the edge $e_{k}=(v_{i}, v_{j})$ to
a non-negative weight $w_{ij}$.
$W=\{w_{ij}|w_{ij}=\psi(e_{k}), e_{k}=(v_{i}, v_{j})\in E\}$ is the weight set.
$G(V,E,W)$ is called a weighted graph.
Referring to the textbook \citep{Balakrishnan2012A},
some properties of a binary graph $G(V,E)$ are briefly introduced as follows:

{\bf Density}: The density of $G(V,E)$ is the ratio of the number of edges
in $E$ to the maximum possible number of edges,
\begin{equation} \label{eq:density}
	density=\frac{2m}{n(n-1)}
\end{equation}
where $n$ and $m$ are the number of nodes and edges in $G$, respectively.
In complex network analysis,
the density has an important impact on other network properties.

{\bf Average shortest path length}:
The shortest path length $l_{ij}$ between nodes $v_{i}$ and $v_{j}$ is
defined as the shortest length of all paths between $v_{i}$ and $v_{j}$.
This length can also be called the distance between the two nodes.
The average shortest path length $L$ of $G(V,E)$ is the average value
of the shortest path lengths between all node pairs:
\begin{equation} \label{eq:ave_len}
	L=\frac{1}{n(n-1)}\sum_{v_{i},v_{j}\in V,i \neq j}l_{ij}
\end{equation}

{\bf Global efficiency}:The global efficiency $E_{global}$ of $G(V,E)$ is as follows:
\begin{equation} \label{eq:glo_eff}
	E_{global}=\frac{1}{n(n-1)}\sum_{v_{i},v_{j}\in V, i\neq j}\frac{1}{l_{ij}}
\end{equation}
where $l_{ij}$ is the shortest path length between nodes $v_{i}$ and $v_{j}$.

{\bf Clustering coefficient}:
For a node $v_{i}$,
$\mathbb{S}_{i}=\{(v_{j},v_{k})|(v_{j},v_{k}), (v_{i}, v_{j})\ and\ (v_{i}, v_{k})\in E\}$ 
represents the edge set between its neighborhoods.
The clustering coefficient $c_{i}$ of the node $v_{i}$ is defined as
the ratio of the number of actual edges between its neighborhoods
and the maximum possible number of edges between them:
\begin{equation} \label{eq:clustering}
	c_{i}=\frac{2|\mathbb{S}_{i}|}{k_{i}(k_{i}-1)}
\end{equation}
where $k_{i}$ is the number of the neighborhoods of $v_{i}$
and $|\mathbb{S}|$ is the number of actual edges between them.
The average clustering coefficient $C$ of $G(V, E)$ is the
average value of the clustering coefficients of all nodes:
\begin{equation} \label{eq:ave_clu}
	C=\frac{1}{n}\sum_{v_{i}\in V}c_{i}
\end{equation}

\subsection{TDA}
\label{TDA}

TDA is a rapidly developing mathematical tool based on algebraic topology,
which can effectively extract the topological information of data.
Please refer to \citep{2009Computational} for more details.

{{\bf Simplicial homology}:
	The simplicial complex is the core object of TDA.
	A $k$-simplex $\tau$ is the convex hull of $k+1$ vertices.
	A $0$-simplex is a vertex, a $1$-simplex is a line segment,
	and a $2$-simplex is a triangle.
	A face of $\tau$ is the convex hull of any subset of the $k+1$ vertices.
	A simplicial complex $K$ is a set of simplexes satisfied:
	(1) All faces of a simplex in $K$ are in $K$.
	(2) The intersection of two simplexes in $K$ is their common face.
	Simplicial homology uses homology groups to describe the
	topological invariants of a simplicial complex.
	The rank of its $k$-dimensional homology group
	is called the $k$-dimensional Betti number $\beta_{k}$
	and represents the number of $k$-dimensional holes ($k$-holes).
	The Betti numbers $\beta_{0}$, $\beta_{1}$, and $\beta_{2}$ represent the
	numbers of connected components, loops, and voids contained in $K$,
	which is the simplification of topological information.
	
	{\bf Persistent homology}:
	Persistent homology was developed to characterize the topological
	information of real data with noise.
	The super-level filtration adapted in this study is defined as follows:
	Given a simplicial complex $K$ and a weighting function
	$\phi$ that maps the simplex $\tau$ in $K$ to a weight $w_{i}$,
	and the weight of $\tau$ must be less than or equal to
	the weights of its faces.
	When the value $\varepsilon$ is chosen,
	all the simplexes with weights greater than $\varepsilon$ form
	a simplicial complex $K_{i}$.
	Moreover, a simplicial complex sequence of $K$:
	$\emptyset = K_{0} \subseteq K_{1} \subseteq ... \subseteq K_{m} = K $
	is obtained by reducing the value $\varepsilon$.
	
	In the filtration,
	a $k$-hole that generates at $\varepsilon_{i_{1}}$ and dies at
	$\epsilon_{i_{2}}$ can be represented by a point
	$(\epsilon_{i_{1}},\epsilon_{i_{2}})\in \mathbb{R}^{2}$.
	All points representing the $k$-holes are drawn on a
	two-dimensional plane,
	defined as a $k$-dimensional persistence diagram ($k$-PD) $D_{k}$,
	which describes the birth and death times of all $k$-holes.
	The $k$-dimensional Betti number sequence of $K$
	can be induced from $D_{k}$,
	which is called the $k$-dimensional Betti curve
	$\beta_{k}(\varepsilon)$ \citep{DONG2021102004}.
	This Betti curve describes the topological invariant $\beta_{k}$
	that persists across multiple scales.
	

\section{Construction of Functional Networks}

\begin{figure*}[ht]
	\vskip 0.2in
	\begin{center}
		\center{\includegraphics[width=\columnwidth]{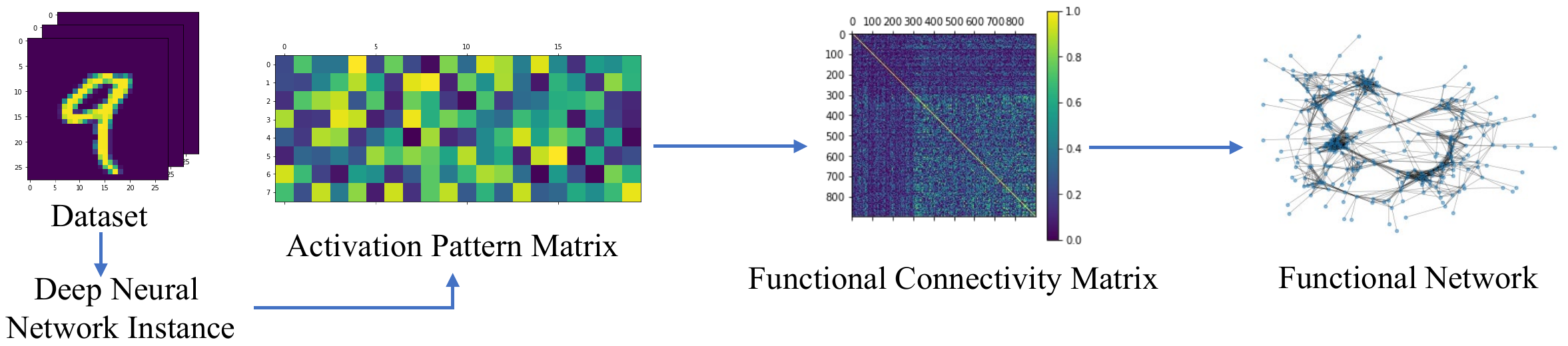}}
		\caption{Flow chart of constructing the functional network for a given deep neural network.}
		\label{fig:flowchart}
	\end{center}
	\vskip -0.2in
\end{figure*}

In this section, we introduce the method for constructing the
functional network of an FCN.    
As shown in Figure \ref{fig:flowchart},
the activation pattern matrix $\mathcal{A}$ is first generated
to construct the functional network for a given FCN.
Then, the functional connectivity matrix $\mathcal{F}$ is obtained  
by calculating the correlation matrix $\mathcal{R}$ according to 
the activation pattern matrix $\mathcal{A}$.
Finally, the functional network $G$ is produced by binarizing $\mathcal{F}$.
The details for generating the functional network are elucidated as follows.

\textbf{Activation Pattern Matrix Generation}
Suppose a trained FCN model $\mathbb{M}$ with $l  (l \geq 1)$ 
hidden layers and $n$ hidden neurons in total 
and a dataset $\mathbb{D}$ with $m$ data are given.
First, data $d_{i} \in \mathbb{D}$ is inputted into
$\mathbb{M}$,
the output value of the $j^{th}$ hidden neuron is denoted as $a_{ij}$,
and an \emph{activation pattern} $(a_{i1},a_{i2},...,a_{in})$
is generated,
which represents the information processing of $\mathbb{M}$ for $d_{i}$.
The \emph{activation pattern matrix} $\mathcal{A}$ is defined by 
considering all the data $d_{i} \in \mathbb{D}  (i=1,2,\dots,m)$:
\begin{equation} \label{eq:ap_matrix}
	\mathcal{A} = (a_{ij})_{m\times n}, 1 \leq i \leq m, 1 \leq j \leq n.
\end{equation}

In the activation pattern matrix $\mathcal{A}$,
the $j^{th}$ column vector is denoted as
$A_j = [a_{1j}, a_{2j}, \cdots, a_{mj}]^T$,
which represents the output of the $j^{th}$ hidden neuron.
If the output vectors $A_i$ and $A_j$ ($i \neq j$) are statistically 
dependent, 
then the $i^{th}$ and $j^{th}$ neurons have functional synergy,
which is called \emph{functional connectivity}.

\textbf{Functional Connectivity Matrix Construction}
The statistical dependency of two hidden neurons is calculated 
based on the activation pattern matrix $\mathcal{A}$.
Many methods can be used to measure the statistical dependency
between two variables,
including Pearson correlation, partial correlation,
and mutual information \citep{Alex2016Fundamentals}.
Compared with Pearson correlation,
the estimation of partial correlation is difficult
\citep{Srikanth2012Estimation}
because of a large number of neurons in $\mathbb{M}$.
In addition, the inaccurate estimation of mutual information 
limits its applications in practice \citep{Daub2004Estimating}.
Consequently, we choose Pearson correlation to measure
the degree of the linear relationship between two neurons
\citep{Heumann2016Association}.
The advantages of the Pearson correlation coefficient $r$ are threefold:
low computational complexity,
ranging from $[-1,1]$,
and the clear significance, that is, 
the closer the absolute value is to 1,
the stronger the linear relationship between two variables is.

To summarize the correlations between all possible pairs of neurons,  
the correlation matrix $\mathcal{R}$ of output values of hidden neurons
is defined by the activation pattern matrix $\mathcal{A}$ as follows:

\begin{definition}[Correlation Matrix]
	\label{def:cor_mtx}
	Given an activation pattern matrix $\mathcal{A}$,
	the Pearson correlation coefficient $r_{ij}$ between the output values 
	of the $i^{th}$ and $j^{th}$ hidden neurons is defined as follows:
	\begin{equation} \label{eq:cor}
		r_{ij}=\frac{\sum_{k=1}^{m}(a_{ki}-\overline{a_{\star i}})(a_{kj}-\overline{a_{\star j}})}{\sqrt{\sum_{k=1}^{m}(a_{ki}-\overline{a_{\star i}})^2} \sqrt{\sum_{k=1}^{m}(a_{kj}-\overline{a_{\star j}})^2}},
	\end{equation}
	where $\overline{a_{\star j}} =\frac{1}{m} \sum_{k=1}^{m}a_{kj}$ is 
	the average output value of the $j^{th}$ hidden neuron.
	Then, the correlation matrix $\mathcal{R}=(r_{ij})_{n \times n}$
	is formed by $r_{ij}$.
\end{definition}

We take the absolute value of $r_{ij}$ as the strength of the
functional connectivity between the $i^{th}$ and $j^{th}$
hidden neurons
and assume that a neuron has no functional connectivity with itself.
Accordingly, the functional connectivity matrix $\mathcal{F}$ 
is defined as follows:
\begin{definition}[Functional Connectivity Matrix] \label{def:fc_mtx}
	Given a correlation matrix $\mathcal{R}=(r_{ij})_{n \times n}$,
	the functional connectivity matrix $\mathcal{F}=(f_{ij})_{n \times n}$
	is defined as follows:
	\begin{equation}\label{eq:fc}
		f_{ij}=\begin{cases}
			|r_{ij}|,&\text{if } i \neq j,\\
			0,&\text{if } i = j;
		\end{cases}
		1 \leq i,j \leq n,
	\end{equation}
	where $f_{ij}$ is the strength of the functional connectivity between
	the $i^{th}$ and $j^{th}$ hidden neurons, $0 \leq f_{ij} \leq1$.
\end{definition}

The functional connectivity matrix $\mathcal{F}$ represents
a weighted complete graph $F(V,E_{f},W_{f})$.
Here, $V=\{v_{1},v_{2},\dots,v_{n}\}$ is the node set,
and the node $v_{j}$ represents the $j^{th}$ hidden
neuron in $\mathbb{M}$.
$E_{f}=\{e_{1},e_{2},\dots,e_{n(n-1)/2}\}$ is the edge set,
and the edge $e_{k}=(v_{i},v_{j})=(v_{j},v_{i})$ represents the functional
connectivity between the $i^{th}$ and $j^{th}$ hidden neurons.
Moreover, a weight function $\psi: E_{f}\rightarrow \mathbb{R}^{+}\cup \{0\}$ is
induced from $F$ that maps the edge $e_{k}=(v_{i},v_{j})$ to a
non-negative weight:
$\psi(e_{k})=w_{k}=f_{ij}$.
$W_{f}=\{w_{k}|\psi(e_{k})=w_{k}, e_{k}\in E_{f}\}$ is the weight set.
$F(V,E_{f},W_{f})$ encodes the statistical dependencies 
between the hidden neurons in $\mathbb{M}$.
However, considering all functional connectivities is inefficient.
By binarizing $F$, the graph's structure is simplified, and 
the functional network is generated. 

\textbf{Functional Network Formation}
In neuroscience, 
network binarization is a common method to construct the functional brain 
network from a weighted complete graph.
We used it to extract the functional network $G(V,E)$ from $F(V,E_{f},W_{f})$.
First, 
the maximum spanning tree $T(V,E_{t},W_{t})$ of $F(V,E_{f},W_{f})$,
which is the spanning tree with a maximum weight of $F$, 
is constructed as the main node-edge structure of 
$G(V,E)$ to ensure its connectivity.
This spanning tree contains all the nodes and $n-1$ edges with the density 
of $2/n$,
which is the ratio of the number of edges in the graph to
that of the corresponding complete graph.
$E_{t}$ is a subset of $E_{f}$,
and $W_{t}=\{w_{k}|\psi(e_{k})=w_{k},e_{k}\in E_{t}\}$ is the weight set.
Then, more edges are required to be added to it
because the tree structure cannot completely depict the interactions 
among neurons.
One of the alternatives is to empirically select a density threshold $d$
for $G(V,E)$, 
which satisfies $2/n \leq d \leq 1$.
This selection means that the functional network $G(V,E)$ should contain 
$[d\times n\times (n-1) /2]$ edges.
We construct $G(V,E)$ as follows:
\begin{definition}[Functional Network]\label{def:func_net}
	Given a density $d$ and a weighted complete graph $F(V,E_{f},W_{f})$,
	let $E_{t}$ be the edge set of its maximum spanning tree.
	The weights of edges in the $E_{f}\setminus E_{t}$ are sorted in a 
	non-ascending order:
	$w'_{1} \geq w'_{2} \geq ... \geq w'_{(n-1)(n-2)/2}$,
	and the edge set of the functional network is defined as
	{\small
		\begin{equation}\label{eq:func_net}
			E=E_{t}\cup \{e_{k}\in E_{f}\ |\ \psi(e_{k})\geq w'_{\left[d\times n\times (n-1)/2\right]-(n-1)}\}.
	\end{equation}}The binary graph $G(V,E)$ is called the functional network.
\end{definition}

The selection of $d$ determines the structure of $G(V,E)$, 
but a unified density selection method does not exist. 
In neuroscience, many studies show that the sparse network can better manifest 
the differences of the functional organization
\citep{Varoquaux2013Cohort,Lv2013Sparse}.
Therefore, we empirically selected multiple small densities for 
the network binarization in practical applications.

In summary, 
the defined functional network $G(V,E)$ depicts the functional interaction
between hidden neurons in the neural network globally, 
regardless of whether physical connectivities exist,
breaking the fixed connection relationships between them.
By adopting the functional network, GTA and TDA can be used to capture 
the functional organization of deep neural networks
and explain and distinguish various models based on their functional differences, 
supported by the results of the following experiments.

\section{Experiments}
In this section, we demonstrate the utility and significance of the 
functional network for FCNs through some experiments.
First, we explore the small-worldness of the functional network,
which is observed on the functional brain network 
in the studies \citep{Young2000Computational,Stam2004Functional}.
Second, we investigate the impact of two commonly used regularization techniques
and explain how they work using GTA and TDA.
Finally, unsupervised clustering is performed according to the 
topological differences between the functional networks 
to demonstrate the effectiveness of TDA in 
evaluating and distinguishing the FCNs.
The Fashion-MNIST \citep{Han2017Fashion}, MNIST \citep{Deng2012MNIST}, and 
CIFAR-10 \citep{Krizhevsky2009Learning} datasets are employed 
in all experiments. 

\subsection{Datasets and Models}
\label{App:DataModel}

The MNIST, Fashion-MNIST, and CIFAR-10 datasets were employed in the experiments.
The contents of the MNIST, Fashion-MNIST, and CIFAR-10 datasets are $28\times 28$
grayscale handwritten digits, $28\times 28$ grayscale fashion products images,
and $32\times 32$ color photographs, respectively
\citep{Deng2012MNIST,Han2017Fashion,Krizhevsky2009Learning}.

We used leaky ReLU activation functions with the negative slope of 0.01 
in the hidden layers
and the Adam optimizer with a learning rate of $3\times 10^{-4}$. 
Each deep neural network was trained for 100 epochs with a batch size of 64.
Tables \ref{tab:small-world-models} and \ref{tab:reg-models} show the
architecture of the deep neural network
trained in the small-world and regularization experiments, respectively.

\begin{table}[t]
	\begin{center}
		\caption{Architectures of the FCNs trained in the small-world experiments.}
		\label{tab:small-world-models}
		\vskip 0.15in
		\begin{center}
			\begin{small}
				\begin{sc}
					\begin{tabular}{lccc}
						\toprule
						\textbf{Dataset} & \textbf{Number of hidden layers} & \textbf{Architecture} \\
						\midrule
						\multirow{2}{*}{MNIST} & {2} &{[300,100],[300,300]}\\  & {3} & {[300,300,100], [300,300,300]}\\ 
						\midrule
						\multirow{2}{*}{Fashion-MNIST} & {2} & {[400,200],[400,400]}\\ & {3} & {[400,400,200],[400,400,400]}\\ 
						\midrule
						\multirow{2}{*}{CIFAR-10} & {2} &{[500,300],[500,500]}\\& {3} & {[500,500,300],[500,500,500]}\\ 
						\bottomrule
					\end{tabular}
				\end{sc}
			\end{small}
		\end{center}
		\vskip -0.1in
		\endgroup
	\end{table}

	\begin{table}[t]
		\caption{Architectures of the FCN trained in the regularization experiments.}
		\label{tab:reg-models}
		\vskip 0.15in
		\begin{center}
			\begin{small}
				\begin{sc}
					\begin{tabular}{lccc}
						\toprule
						\textbf{Dataset} & \textbf{Number of hidden layers} & \textbf{Architecture} \\
						\midrule
						\multirow{2}{*}{Fashion-MNIST} & {2} & {[400,200],[400,400]}\\ & {3} & {[400,400,200],[400,400,400]}\\ 
						\midrule
						\multirow{2}{*}{MNIST} & {2} &{[300,100]}\\  & {3} & {[300,300,300]}\\ 
						\midrule
						\multirow{2}{*}{CIFAR-10} & {2} &{[500,300]}\\& {3} & {[500,500,300]}\\ 
						\bottomrule
					\end{tabular}
				\end{sc}
			\end{small}
		\end{center}
		\vskip -0.1in
	\end{table}
	

\subsection{Small-World Experiments}

\begin{figure*}[ht]
	\center{\includegraphics[width=\columnwidth]  {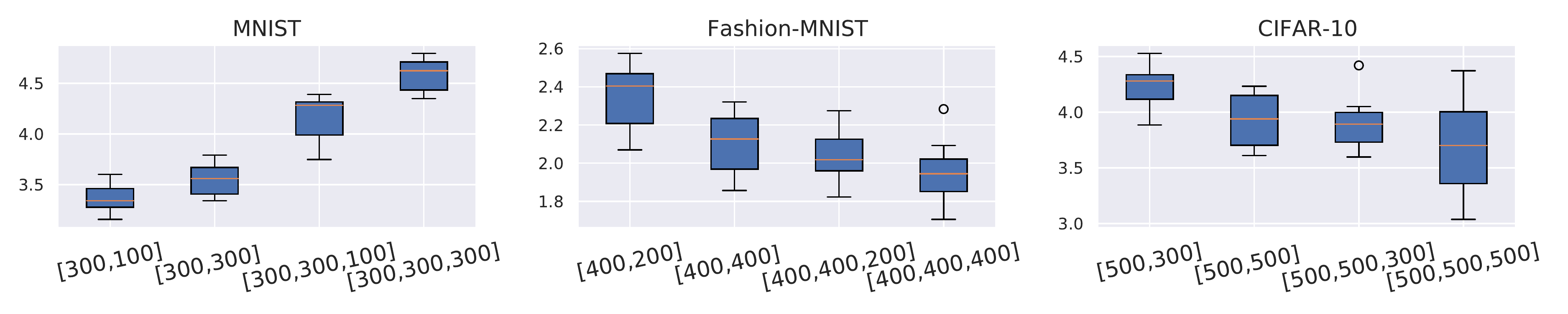}}
	\caption{\label{fig:Small-Worldness}
		Box plots for the small-world coefficients of the functional networks 
		for the FCNs with different architectures.
	}
\end{figure*}

The network with a large average clustering coefficient
and a small average shortest path length is called a small-world network.
This property is called the small-worldness, 
which can be measured by the small-world coefficient $\sigma$
\citep{Humphries2008network}:
\begin{equation} \label{eq:small_world}
	\sigma=\frac{C_{real}/C_{random}}{L_{real}/{L_{random}}},
\end{equation}
where $C_{real}$ and $C_{random}$ mean the average clustering coefficient
of the real network and its equivalent random network, respectively;
$L_{real}$ and $L_{random}$ mean the average shortest path length
of the real network and its equivalent random network, respectively.
If $\sigma$ is greater than 1,
then the real network is deemed a small-world network.
The larger $\sigma$ is,
the more significant the small-worldness is.

Previous studies \citep{Young2000Computational,Stam2004Functional}
suggested that the functional brain network is a
small-world network
and any two brain regions only have a small number 
of intermediate steps to connect.
This functional organization improves the efficiency of global information 
transmission in the brain 
and reflects the optimal organization pattern for information processing 
\citep{Strogatz2001Exploring,Silke2002In,Egu2005Scale}.
Analogously, a worthy question arises to explore the existence
of the small-worldness in the functional network of deep neural networks.
To study it, we used the FCNs trained on three datasets.
For each dataset,
we trained four groups of FCNs with different architectures,
as shown in Table \ref{tab:small-world-models} in 
Section \ref{App:DataModel}.
Each group includes 10 FCNs with only the initial values diverse.
Then, we constructed their functional networks with a density of $2.5\%$,
calculated the small-world coefficients $\sigma$,
and illustrated the results in Figure \ref{fig:Small-Worldness}.

For the FCNs trained on the MNIST and CIFAR-10 datasets,
the small-world coefficients of all functional networks are between 3.0 
and 5.0,
whereas for the FCNs trained on the Fashion-MNIST dataset, 
the values are between 1.6 and 2.6.
The small-world coefficients are all greater than 1.0 for the trained FCNs.
This result suggests that the functional network of FCNs is a small-world network,
which is general for FCNs with different initial values, architectures, and
training datasets.
Meanwhile, Figure \ref{fig:Small-Worldness} shows that the proportional relation
between the small-world coefficients and the width and depth of network
architectures does not exist.

The small-world experiments illustrate that FCNs have a functional network 
that is similar to and efficient as the functional network 
examined in the brain. 
The brain-like functional organization of deep neural networks enhances their
information transmission and processing capability, 
ensuring optimal model performance.

\subsection{Regularization Experiments}
Deep neural networks have a high capacity and are prone to over-fit.
Therefore, a number of regularization strategies
\citep{moradi2020survey}
have been developed to improve generalization,
such as batch normalization \citep{Ioffe2015Batch} and dropout
\citep{Srivastava2014Dropout}.
Previous studies showed that dropout increases the robustness of 
deep neural networks 
\citep{El2017On,Park2017Analysis},
whereas batch normalization reduces it \citep{benz2021Batch}.
Moreover, when batch normalization and dropout are combined practically, 
model performance degrades \citep{Ioffe2015Batch,Li2019Understanding}.
Our experiments investigate the mechanisms of batch normalization and dropout
and explain the results mentioned above.

In the experiments, 
we trained FCNs with different architectures on three datasets,
as shown in Table \ref{tab:reg-models} in Section \ref{App:DataModel}.
For each architecture, 
we trained three groups of FCNs, 
where each group includes 20 models with only the initial values diverse. 
The models in the first group (\emph{vanilla group}) 
were trained without regularization,
whereas the models in the second (\emph{dropout group}) 
and third groups (\emph{BatchNorm group})
were trained with dropout and batch normalization, respectively.
The dropout rate was set to 50\%.

We used GTA and TDA to explore how dropout and batch normalization 
affect the functional network of FCNs. 
The effects of regularization strategies on network functional interaction
patterns are reflected in the graph theoretical and topological properties, 
which explains their mechanisms. 
To show that topological features from TDA characterize FCNs,
we employed hierarchical clustering, an unsupervised method, 
to identify the models through TDA features,
compared with the clustering results using test accuracies.

\subsubsection{GTA}

\begin{figure*}[bt]
	\center{\includegraphics[width=\columnwidth]  {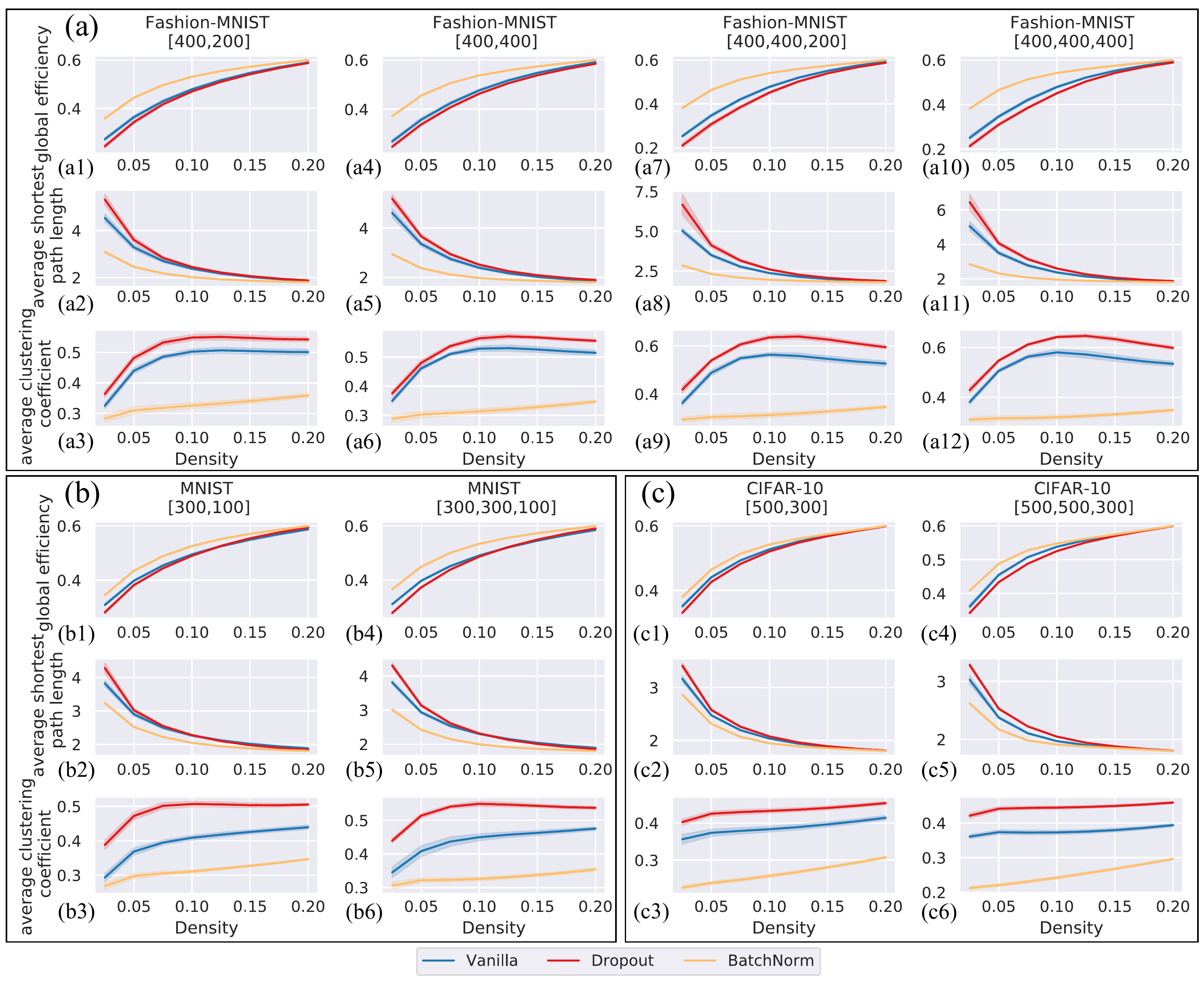}}
	\caption{\label{fig:GTA}
		Graph theoretical properties of the functional networks with different
		densities for the FCNs trained on the (a) Fashion-MNIST, 
		(b) MNIST, and (c) CIFAR-10 datasets.
		The blue, red, and orange curves represent the corresponding average index with 	 
		the error bars of the functional networks on the first,
		second, and third groups, respectively.}
\end{figure*}
%
%
For each group of the trained FCNs,
we constructed and analyzed their functional networks using GTA.
First, an appropriate density for the functional network construction
should be selected.
A high density introduces too much noise, 
whereas a low density makes important connectivities removed. 
As a result, for FCN,
we constructed the functional network sequence by setting the density
ranging from 2.5\% to 20\% with a 2.5\% increment.
Then, we characterized the FCNs by the global efficiency,
average shortest path length, and average clustering coefficient
of their functional networks. 
Specifically, the clustering coefficient of a node measures the proportion
of edges between its neighborhood, divided by the maximum
possible number of edges.
In complex brain network analysis, 
the former two measure functional integration and information
transmission of the network.
The latter assesses the topological redundancy, 
local fault tolerance, and functional specialization
\citep{Alex2016Fundamentals}.

As shown in Figure \ref{fig:GTA},
the standard deviations of the graph theoretical properties 
are small for the FCNs in the same group,
which suggests that the graph theoretical properties are
stable to the initial values.
Moreover, at the same density,
the global efficiency of the BatchNorm group is over that of the vanilla group, 
whereas that of the dropout group is below it.
Meanwhile, for the average shortest path length and average clustering coefficient,
the corresponding relation of quantity between them is opposite:
the values of the dropout group are larger than those of the vanilla group,
and those of the BatchNorm group are less.
The graph theoretical differences between the functional networks
in various groups show that the regularization techniques 
have different impacts on FCNs.
The models with batch normalization have higher global information 
transmission capability and more rapid, integrated, and efficient
communication between neurons, 
which improves network performance.
However, the improvement of efficiency comes at the cost of a decrease
in the clustering coefficient.
The clustering coefficient measures the network fault tolerance, 
and a highly clustered network is resilient to random attacks 
\citep{Alex2016Fundamentals}. 
The decrease in the fault tolerance leads to a decline in 
adversarial robustness.

In contrast to batch normalization, 
dropout raises the average clustering coefficient and average shortest path length 
while lowering the global efficiency. 
The high average clustering coefficient demonstrates that numerous small subgraphs
with tight internal integration exist in the functional network.
Moreover, the neurons in a subgraph encode comparable features,
which can be considered a functional group.
The functional groups facilitate functional specialization within the network,
which contributes to fast and efficient information processing \citep{Ringo1994Time},
and increase the network redundancy and fault tolerance.
Therefore, the robustness and model performance are improved.

In conclusion, batch normalization and dropout have different mechanisms:
(1) Batch normalization enhances model performance by increasing the 
global efficiency of neural networks
but reduces adversarial robustness by lowering the fault tolerance. 
(2) Dropout facilitates functional specialization and fault tolerance
by increasing functional groups, 
which improves the generalization ability and robustness of neural networks. 
According to this conclusion, we can explain the decline in network performance
when dropout and batch normalization are combined practically.

\subsubsection{TDA and Clustering Experiments}

\begin{figure*}[t]
	\center{\includegraphics[width=\columnwidth] {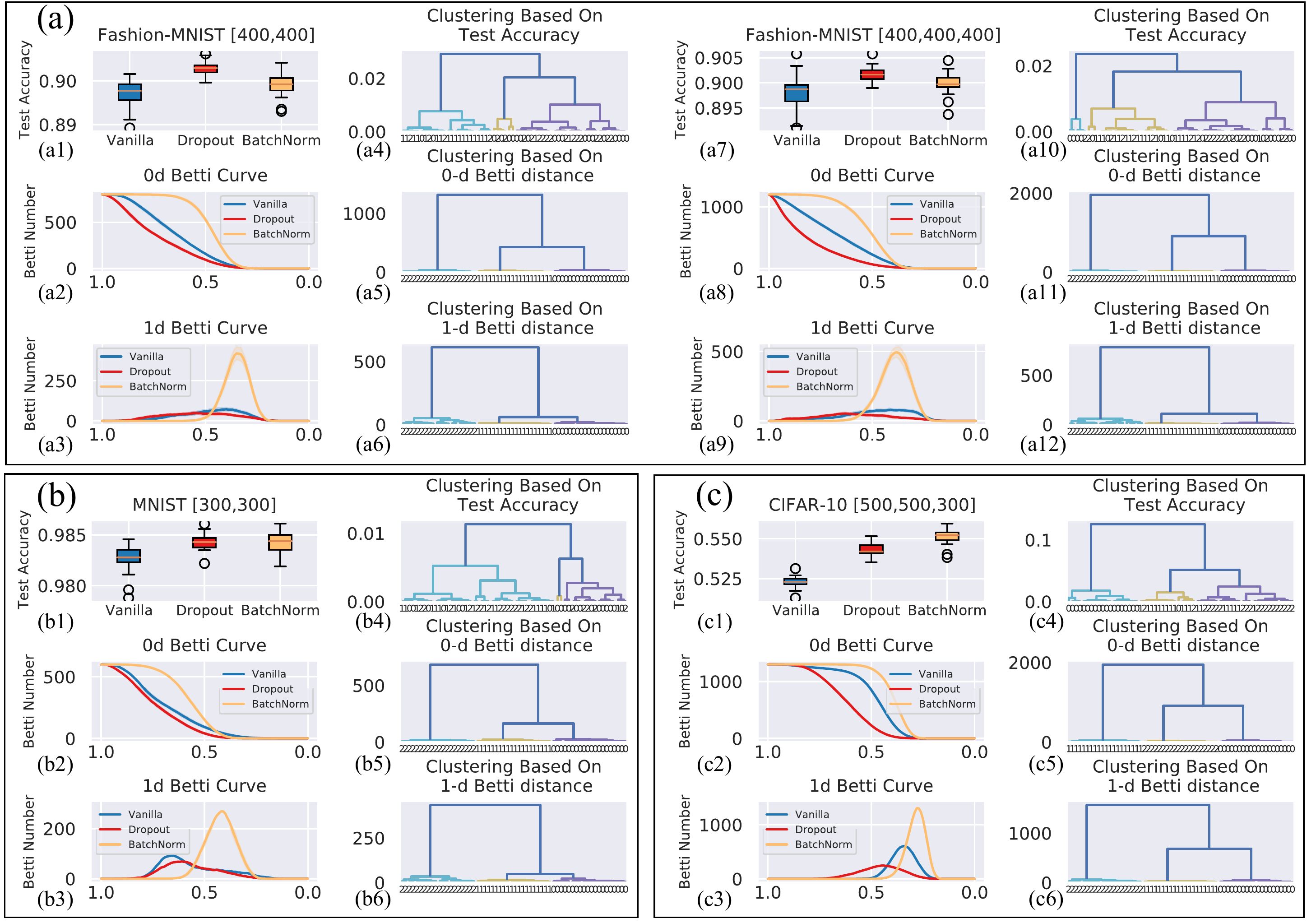}}
	\caption{\label{fig:TDA}
		Direct outcome of TDA for the FCNs trained on the (a) Fashion-MNIST,
		(b) MNIST, and (c) CIFAR-10 datasets:
		a1, a7, b1, and c1: the box  plots of the test accuracies for the FCNs;
		a2, a3, a8, a9, b2, b3, c2, and c3: the average zero- and one-dimensional 
		Betti curves with the error bars of the functional networks;
		a4-a6, a10-a11, b4-b6, and c4-c6:
		the dendrograms of hierarchical clustering by
		test accuracies, and zero- and one-dimensional Betti distances, 
		where 0, 1, and 2 denote the FCNs in the vanilla, dropout, and 
		BatchNorm groups, respectively.}
\end{figure*}

Compared with GTA, TDA depicts the higher-dimensional interactions 
at different resolutions without the density selection in GTA
and is more robust to noise.
In TDA, a network is modeled as a simplicial complex $K$,
which is a set of simplexes $\tau$.
A $k$-simplex $\tau_{i}$ is the convex hull of $k+1$ vertices,
denoted as $\tau_{i}=[v_{i0},v_{i1},\cdots,v_{ik}]$.
A face of $\tau_{i}$ is the convex hull of its vertice subset.

Previous studies \citep{Rieck2019Neural,Watanabe2021Topological} showed that 
the complexity of FCNs could be measured by their zero- 
and one-dimensional structural topological features.
Meanwhile, TDA is also used to capture the topological differences between 
functional brain networks to identify and classify 
various types of brains
\citep{Billings2021Simplicial}.
In this work, we applied TDA to obtain the zero-
and one-dimensional Betti number curve of the functional
network to explain the mechanisms of dropout and batch normalization.
To show that TDA characterizes deep neural networks,
the clustering experiments were performed to distinguish the FCNs trained with
different regularizations according to their functional topological features.

First, we modeled a functional network as a weighted simplicial complex $K$.
Given a weighted graph $F(V, E_{f}, W_{f})$ with the weight function $\psi$,
we can define a simplicial complex
$K=\{[v_{i0}, v_{i1},..., v_{ik}]|v_{i0}, v_{i1},..., v_{ik}\in V
{\ and\ } 0 \leq k < n\}$ 
with a weight function $\phi: K\to \mathbb{R}$, \textit{i.e.},
\begin{equation}
	\phi(\tau_{i})=\begin{cases}
		1,&\text{if }\tau_{i} = [v_{i0}],\\
		\psi((v_{i0},v_{i1})), &\text{if }\tau_{i} = [v_{i0},v_{i1}],\\
		\mathop{\min}_{\kappa \subset \tau_{i}} \phi(\kappa), & otherwise,
	\end{cases}\label{eq:weight_fun}
\end{equation}
where $\psi((v_{i0},v_{i1}))$ is the weight of the edge $(v_{i0},v_{i1})$ in $F$,
and $\kappa$ represents any face of the simplex $\tau_{i}$.
Then, the super-level filtration was adapted to get the $k$-dimensional Betti number
curve $\beta_{k}(\varepsilon)$ of $K$, where $\varepsilon$ represents
the filtration threshold \citep{DONG2021102004}. 

For $F(V,E_{f},W_{f})$,
the nodes in $V$ are modeled as 0-simplexes with the weights of 1 in $K$.
The functional connectivities in $E_{f}$ are modeled as $1$-simplexes with
the weights of corresponding functional connectivity strength.
Moreover, the $k$-cliques ($k\geq 2$) in $F(V,E_{f},W_{f})$ are
modeled as k-simplexes with weights equal to the minimum
weight of their faces.

We analyzed the FCNs with architectures [400, 400] and [400, 400, 400] trained 
on the Fashion-MNIST dataset,
the FCNs with [300, 300] trained on the MNIST dataset,
and the FCNs with [500, 500, 300] trained on the CIFAR-10 dataset.
For the FCN $\mathbb{M}_{i}$, 
we constructed the corresponding weighted simplicial complex $K^{i}$
and obtained the zero- and one-dimensional Betti number 
curves $\beta_{0}^{i}(\epsilon)$ and $\beta_{1}^{i}(\epsilon)$
by filtering $K^{i}$ from 1 to 0.
The Betti numbers $\beta_{0}^{i}$ and $\beta_{1}^{i}$ represent the
numbers of connected components and loops contained in $K^{i}$, respectively.
Moreover, the zero- and one-dimensional Betti distances between
functional networks were calculated as follows to measure
the functional topological differences between the FCNs:
\begin{equation} \label{eq:betti_dis}
	d_{k}(K^{i},K^{j})=\int_{\varepsilon} |\beta_{k}^{i}(x)-\beta_{k}^{j}(x)|^{2} dx. 
\end{equation}
where $k=0, 1$.
Finally, we clustered the networks hierarchically according to their
test accuracies, 
and zero- and one-dimensional Betti distances. 
We illustrated the clustering results in the dendrograms.

Figure \ref{fig:TDA} (a1, a7, b1, and c1)
shows that the median test accuracies of the FCNs with regularization 
are higher than those of the vanilla FCNs.
Figure \ref{fig:TDA} (a2, a3, a8, a9, b2, b3, c2, and c3) displays the 
average zero- and one-dimensional Betti curves.
The small standard deviations of the Betti curves imply that the Betti numbers
of functional networks are stable to the initial values.

For the simplicial complexes at the same threshold value $\epsilon$,
average $\beta_{0}$ in the BatchNorm group is the largest,
followed by that in the vanilla and dropout groups,
whereas average $\beta_{0}$ in the dropout group is the smallest.
This result indicates that the functional networks in the dropout group have the least 
number of connected components at the same threshold.
That is, the FCNs with dropout possess several functional groups,
in which the neurons encode similar features and thus have
strong functional connectivities.
Therefore, the connected components in the same functional group merge early,
causing a small $\beta_{0}$.
On the contrary, the FCNs with batch normalization have few and weak functional groups,
leading to a large $\beta_{0}$.

Compared with the peak of one-dimensional Betti curves for the vanilla group, 
that for the BatchNorm group is significantly higher,
whereas the peak for the dropout group is lower.
The maximum $\beta_{1}$ represents the maximum number of loops that occur
in the filtration of the simplicial complex.
The combinational effects of numerous neurons in deep neural networks can be
revealed through one-dimensional topological features 
\citep{Watanabe2021Topological}. 
The loops might attribute to the feature coding in deep neural networks. 
The previous study \citep{benz2021Batch} showed that batch normalization 
allows the utilization of more useful features to increase accuracy. 
The highest peak of the one-dimensional Betti curves in the BatchNorm group
may suggest that batch normalization potentially improves the coding capability
of neural networks by increasing the number of functional loops.
Furthermore, $\beta_{0}$ and the maximum $\beta_{1}$ of 
the functional networks in the dropout and BatchNorm 
groups change in the opposite tendency.
This result implies that dropout and batch normalization have opposing impacts on
zero- and one-dimensional topological features, 
which is in accordance with the observation in GTA. 

As shown in Figure \ref{fig:TDA} (a4-a6, a10-a12, b4-b6, and c4-c6),
the FCNs with various regularizations can be correctly clustered according to
the zero- and one-dimensional Betti distances
while being incorrectly clustered simply using the test accuracies. 
Although regularization methods enhance network performance,
the improvements in test accuracies are insufficient to distinguish
regularization techniques.
This finding indicates that, compared with test accuracies, 
TDA indexes can better evaluate and distinguish deep neural networks.

Moreover, the best clustering index is the zero-dimensional Betti distance
because it produces a large distance between clusters and
a small distance within a cluster. 
Although the FCNs can be correctly clustered according to the one-dimensional 
Betti distances, 
the distances between the vanilla and dropout groups are
comparatively small.
This finding reflects that batch normalization has a greater impact on the one-dimensional
topological structures of the functional network than dropout.
The findings also suggest the potential of the functional network and TDA for 
extracting functional topological features to 
explain, evaluate, and distinguish deep neural networks.

\section{Conclusion}

In this work, we propose the functional network as a novel
framework for interpretability of deep neural networks.
We show that the functional network of FCNs is a small-world network, 
similar to the brain functional network,
suggesting that the two have a similar functional organization.
Batch normalization enhances model performance by 
increasing the global efficiency and the number of functional loops
but reduces adversarial robustness by lowering the fault tolerance.
Dropout enhances the functional specialization and fault tolerance in models
by increasing the number of functional groups and network redundancy, 
improving the generalization ability and robustness of neural networks.
Additionally, the models with different regularizations were clustered correctly 
according to their functional topological differences,
reflecting that topological features based on TDA characterize the FCNs.

In this work, Pearson correlation is used as the measure 
of statistical dependency between neural activities of neurons.
In future work, we will choose other methods to measure functional
connectivity.
Another interesting avenue is to study the similarities
and differences of coding mechanisms between the brain and the
deep neural network from the perspective of the functional network,
which will promote the research of brain-inspired intelligence.




\newpage

\appendix



\noindent

\vskip 0.2in
\bibliography{refs}

\begin{thebibliography}{57}
\providecommand{\natexlab}[1]{#1}
\providecommand{\url}[1]{\texttt{#1}}
\expandafter\ifx\csname urlstyle\endcsname\relax
  \providecommand{\doi}[1]{doi: #1}\else
  \providecommand{\doi}{doi: \begingroup \urlstyle{rm}\Url}\fi

\bibitem[Akula et~al.(2020)Akula, Wang, and Zhu]{DBLP:conf/aaai/AkulaWZ20}
A.~R. Akula, S.~Wang, and S.~Zhu.
\newblock Cocox: Generating conceptual and counterfactual explanations via
  fault-lines.
\newblock In \emph{The Thirty-Fourth {AAAI} Conference on Artificial
  Intelligence, {AAAI} 2020, The Thirty-Second Innovative Applications of
  Artificial Intelligence Conference, {IAAI} 2020, The Tenth {AAAI} Symposium
  on Educational Advances in Artificial Intelligence, {EAAI} 2020, New York,
  NY, USA, February 7-12, 2020}, pages 2594--2601. {AAAI} Press, 2020.
\newblock URL \url{https://aaai.org/ojs/index.php/AAAI/article/view/5643}.

\bibitem[Balakrishnan and Ranganathan(2012)]{Balakrishnan2012A}
R.~Balakrishnan and K.~Ranganathan.
\newblock \emph{A Textbook of Graph Theory}.
\newblock Springer New York, New York, NY, 2012.
\newblock ISBN 978-1-4614-4529-6.
\newblock \doi{10.1007/978-1-4614-4529-6}.
\newblock URL \url{https://doi.org/10.1007/978-1-4614-4529-6}.

\bibitem[Bassett et~al.(2011)Bassett, Wymbs, Porter, Mucha, Carlson, and
  Grafton]{Bassett2011Dynamic}
D.~S. Bassett, N.~F. Wymbs, M.~A. Porter, P.~J. Mucha, J.~M. Carlson, and S.~T.
  Grafton.
\newblock Dynamic reconfiguration of human brain networks during learning.
\newblock \emph{Proceedings of the National Academy of Sciences}, 108\penalty0
  (18):\penalty0 7641--7646, 2011.
\newblock ISSN 0027-8424.
\newblock \doi{10.1073/pnas.1018985108}.
\newblock URL \url{https://www.pnas.org/content/108/18/7641}.

\bibitem[Bassett et~al.(2015)Bassett, Yang, Wymbs, and
  Grafton]{Bassett2015Learning}
D.~S. Bassett, M.~Yang, N.~F. Wymbs, and S.~T. Grafton.
\newblock Learning-induced autonomy of sensorimotor systems.
\newblock \emph{Nature Neuroscience}, 18\penalty0 (5):\penalty0 744--751, 2015.
\newblock \doi{10.1038/nn.3993}.

\bibitem[Bau et~al.(2017)Bau, Zhou, Khosla, Oliva, and
  Torralba]{Bau2017Network}
D.~Bau, B.~Zhou, A.~Khosla, A.~Oliva, and A.~Torralba.
\newblock Network dissection: Quantifying interpretability of deep visual
  representations.
\newblock In \emph{2017 IEEE Conference on Computer Vision and Pattern
  Recognition (CVPR)}, pages 3319--3327, 2017.
\newblock \doi{10.1109/CVPR.2017.354}.

\bibitem[Beaty et~al.(2018)Beaty, Kenett, Christensen, Rosenberg, Benedek,
  Chen, Fink, Qiu, Kwapil, Kane, and Silvia]{Beaty2018Robust}
R.~E. Beaty, Y.~N. Kenett, A.~P. Christensen, M.~D. Rosenberg, M.~Benedek,
  Q.~Chen, A.~Fink, J.~Qiu, T.~R. Kwapil, M.~J. Kane, and P.~J. Silvia.
\newblock Robust prediction of individual creative ability from brain
  functional connectivity.
\newblock \emph{Proceedings of the National Academy of Sciences}, 115\penalty0
  (5):\penalty0 1087--1092, 2018.
\newblock ISSN 0027-8424.
\newblock \doi{10.1073/pnas.1713532115}.
\newblock URL \url{https://www.pnas.org/content/115/5/1087}.

\bibitem[Benz et~al.(2021)Benz, Zhang, and Kweon]{benz2021Batch}
P.~Benz, C.~Zhang, and I.~S. Kweon.
\newblock Batch normalization increases adversarial vulnerability and decreases
  adversarial transferability: A non-robust feature perspective.
\newblock In \emph{Proceedings of the IEEE/CVF International Conference on
  Computer Vision (ICCV)}, pages 7818--7827, October 2021.

\bibitem[Bi and Zhou(2020)]{Bi2020Understanding}
Z.~Bi and C.~Zhou.
\newblock Understanding the computation of time using neural network models.
\newblock \emph{Proceedings of the National Academy of Sciences}, 117\penalty0
  (19):\penalty0 10530--10540, 2020.
\newblock ISSN 0027-8424.
\newblock \doi{10.1073/pnas.1921609117}.
\newblock URL \url{https://www.pnas.org/content/117/19/10530}.

\bibitem[Bien and Tibshirani(2011)]{10.1214/11-AOAS495}
J.~Bien and R.~Tibshirani.
\newblock {Prototype selection for interpretable classification}.
\newblock \emph{The Annals of Applied Statistics}, 5\penalty0 (4):\penalty0
  2403 -- 2424, 2011.
\newblock \doi{10.1214/11-AOAS495}.
\newblock URL \url{https://doi.org/10.1214/11-AOAS495}.

\bibitem[Billings et~al.(2021)Billings, Saggar, Hlinka, Keilholz, and
  Petri]{Billings2021Simplicial}
J.~Billings, M.~Saggar, J.~Hlinka, S.~Keilholz, and G.~Petri.
\newblock {Simplicial and topological descriptions of human brain dynamics}.
\newblock \emph{Network Neuroscience}, 5\penalty0 (2):\penalty0 549--568, 06
  2021.
\newblock ISSN 2472-1751.
\newblock \doi{10.1162/netn_a_00190}.
\newblock URL \url{https://doi.org/10.1162/netn\_a\_00190}.

\bibitem[Boz(2002)]{Boz2002Extracting}
O.~Boz.
\newblock Extracting decision trees from trained neural networks.
\newblock KDD '02, page 456–461, New York, NY, USA, 2002. Association for
  Computing Machinery.
\newblock ISBN 158113567X.
\newblock \doi{10.1145/775047.775113}.
\newblock URL \url{https://doi.org/10.1145/775047.775113}.

\bibitem[Cadieu et~al.(2014)Cadieu, Hong, Yamins, Pinto, Ardila, Solomon,
  Majaj, Dicarlo, and Bethge]{Cadieu2014Deep}
C.~F. Cadieu, H.~Hong, D.~Yamins, N.~Pinto, D.~Ardila, E.~A. Solomon, N.~J.
  Majaj, J.~J. Dicarlo, and M.~Bethge.
\newblock Deep neural networks rival the representation of primate it cortex
  for core visual object recognition.
\newblock \emph{Plos Computational Biology}, 10\penalty0 (12):\penalty0
  e1003963, 2014.
\newblock \doi{10.1371/journal.pcbi.1003963}.

\bibitem[Dabaghian et~al.(2012)Dabaghian, Mémoli, Frank, and
  Carlsson]{Dabaghian2012A}
Y.~Dabaghian, F.~Mémoli, L.~Frank, and G.~Carlsson.
\newblock A topological paradigm for hippocampal spatial map formation using
  persistent homology.
\newblock \emph{PLoS computational biology}, 8\penalty0 (8):\penalty0 e1002581,
  2012.
\newblock ISSN 1553-734X.
\newblock \doi{10.1371/journal.pcbi.1002581}.
\newblock URL \url{https://europepmc.org/articles/PMC3415417}.

\bibitem[Dalvi et~al.(2019)Dalvi, Durrani, Sajjad, Belinkov, Bau, and
  Glass]{Dalvi2019What}
F.~Dalvi, N.~Durrani, H.~Sajjad, Y.~Belinkov, A.~Bau, and J.~Glass.
\newblock What is one grain of sand in the desert? analyzing individual neurons
  in deep nlp models.
\newblock \emph{Proceedings of the AAAI Conference on Artificial Intelligence},
  33\penalty0 (01):\penalty0 6309--6317, Jul. 2019.
\newblock \doi{10.1609/aaai.v33i01.33016309}.
\newblock URL \url{https://ojs.aaai.org/index.php/AAAI/article/view/4592}.

\bibitem[Daub et~al.(2004)Daub, Steuer, Selbig, and Kloska]{Daub2004Estimating}
C.~O. Daub, R.~Steuer, J.~Selbig, and S.~Kloska.
\newblock Estimating mutual information using b-spline functions--an improved
  similarity measure for analysing gene expression data.
\newblock \emph{BMC bioinformatics}, 5:\penalty0 118, August 2004.
\newblock ISSN 1471-2105.
\newblock \doi{10.1186/1471-2105-5-118}.
\newblock URL \url{https://europepmc.org/articles/PMC516800}.

\bibitem[Deng(2012)]{Deng2012MNIST}
L.~Deng.
\newblock The mnist database of handwritten digit images for machine learning
  research.
\newblock \emph{IEEE Signal Processing Magazine}, 29\penalty0 (6):\penalty0
  141--142, 2012.

\bibitem[Dodel et~al.(2002)Dodel, Herrmann, and Geisel]{Silke2002In}
S.~Dodel, J.~Herrmann, and T.~Geisel.
\newblock Functional connectivity by cross-correlation clustering.
\newblock \emph{Neurocomputing}, 44-46:\penalty0 1065--1070, 2002.
\newblock ISSN 0925-2312.
\newblock \doi{https://doi.org/10.1016/S0925-2312(02)00416-2}.
\newblock URL
  \url{https://www.sciencedirect.com/science/article/pii/S0925231202004162}.
\newblock Computational Neuroscience Trends in Research 2002.

\bibitem[Dong et~al.(2021)Dong, Pu, and Lin]{DONG2021102004}
Z.~Dong, J.~Pu, and H.~Lin.
\newblock Multiscale persistent topological descriptor for porous structure
  retrieval.
\newblock \emph{Computer Aided Geometric Design}, 88:\penalty0 102004, 2021.
\newblock ISSN 0167-8396.
\newblock \doi{https://doi.org/10.1016/j.cagd.2021.102004}.
\newblock URL
  \url{https://www.sciencedirect.com/science/article/pii/S0167839621000492}.

\bibitem[Edelsbrunner and Harer(2009)]{2009Computational}
H.~Edelsbrunner and J.~Harer.
\newblock \emph{Computational Topology: An Introduction}.
\newblock American Mathematical Society, 2009.

\bibitem[Egu\'{\i}luz et~al.(2005)Egu\'{\i}luz, Chialvo, Cecchi, Baliki, and
  Apkarian]{Egu2005Scale}
V.~M. Egu\'{\i}luz, D.~R. Chialvo, G.~A. Cecchi, M.~Baliki, and A.~V. Apkarian.
\newblock Scale-free brain functional networks.
\newblock \emph{Phys. Rev. Lett.}, 94:\penalty0 018102, Jan 2005.
\newblock \doi{10.1103/PhysRevLett.94.018102}.
\newblock URL \url{https://link.aps.org/doi/10.1103/PhysRevLett.94.018102}.

\bibitem[El~Mhamdi et~al.(2017)El~Mhamdi, Guerraoui, and Rouault]{El2017On}
E.~M. El~Mhamdi, R.~Guerraoui, and S.~Rouault.
\newblock On the robustness of a neural network.
\newblock In \emph{2017 IEEE 36th Symposium on Reliable Distributed Systems
  (SRDS)}, pages 84--93, 2017.
\newblock \doi{10.1109/SRDS.2017.21}.

\bibitem[Fornito et~al.(2016)Fornito, Zalesky, and
  Bullmore]{Alex2016Fundamentals}
A.~Fornito, A.~Zalesky, and E.~T. Bullmore.
\newblock \emph{Fundamentals of Brain Network Analysis}.
\newblock Academic Press, San Diego, 2016.
\newblock ISBN 978-0-12-407908-3.
\newblock \doi{10.1016/C2012-0-06036-X}.
\newblock URL \url{https://www.sciencedirect.com/book/9780124079083}.

\bibitem[Güçlü and van Gerven(2015)]{Umut2015Deep}
U.~Güçlü and M.~A.~J. van Gerven.
\newblock Deep neural networks reveal a gradient in the complexity of neural
  representations across the ventral stream.
\newblock \emph{Journal of Neuroscience}, 35\penalty0 (27):\penalty0 10005 --
  10014, 2015.
\newblock ISSN 02706474.
\newblock \doi{10.1523/JNEUROSCI.5023-14.2015}.
\newblock URL \url{https://www.jneurosci.org/content/35/27/10005}.

\bibitem[Heumann et~al.(2016)Heumann, Schomaker, and
  Shalabh]{Heumann2016Association}
C.~Heumann, M.~Schomaker, and Shalabh.
\newblock \emph{Association of Two Variables}, pages 67--94.
\newblock Springer International Publishing, Cham, 2016.
\newblock ISBN 978-3-319-46162-5.
\newblock \doi{10.1007/978-3-319-46162-5_4}.
\newblock URL \url{https://doi.org/10.1007/978-3-319-46162-5_4}.

\bibitem[Humphries and Gurney(2008)]{Humphries2008network}
M.~D. Humphries and K.~Gurney.
\newblock Network 'small-world-ness': a quantitative method for determining
  canonical network equivalence.
\newblock \emph{PLoS ONE}, 3\penalty0 (4), 2008.
\newblock \doi{10.1371/journal.pone.0002051}.

\bibitem[Ioffe and Szegedy(2015)]{Ioffe2015Batch}
S.~Ioffe and C.~Szegedy.
\newblock Batch normalization: Accelerating deep network training by reducing
  internal covariate shift.
\newblock In F.~Bach and D.~Blei, editors, \emph{Proceedings of the 32nd
  International Conference on Machine Learning}, volume~37 of \emph{Proceedings
  of Machine Learning Research}, pages 448--456, Lille, France, 07--09 Jul
  2015. PMLR.
\newblock URL \url{https://proceedings.mlr.press/v37/ioffe15.html}.

\bibitem[Kim et~al.(2014)Kim, Rudin, and Shah]{10.5555/2969033.2969045}
B.~Kim, C.~Rudin, and J.~Shah.
\newblock The bayesian case model: A generative approach for case-based
  reasoning and prototype classification.
\newblock In \emph{Proceedings of the 27th International Conference on Neural
  Information Processing Systems - Volume 2}, NIPS'14, page 1952–1960,
  Cambridge, MA, USA, 2014. MIT Press.

\bibitem[Krizhevsky(2009)]{Krizhevsky2009Learning}
A.~Krizhevsky.
\newblock Learning multiple layers of features from tiny images.
\newblock Technical report, 2009.

\bibitem[Li et~al.(2019)Li, Chen, Hu, and Yang]{Li2019Understanding}
X.~Li, S.~Chen, X.~Hu, and J.~Yang.
\newblock Understanding the disharmony between dropout and batch normalization
  by variance shift.
\newblock In \emph{2019 IEEE/CVF Conference on Computer Vision and Pattern
  Recognition (CVPR)}, pages 2677--2685, 2019.
\newblock \doi{10.1109/CVPR.2019.00279}.

\bibitem[Lv et~al.(2013)Lv, Li, Zhu, Jiang, Zhang, Hu, Zhang, Guo, and
  Liu]{Lv2013Sparse}
J.~Lv, X.~Li, D.~Zhu, X.~Jiang, X.~Zhang, X.~Hu, T.~Zhang, L.~Guo, and T.~Liu.
\newblock Sparse representation of group-wise fmri signals.
\newblock In K.~Mori, I.~Sakuma, Y.~Sato, C.~Barillot, and N.~Navab, editors,
  \emph{Medical Image Computing and Computer-Assisted Intervention -- MICCAI
  2013}, pages 608--616, Berlin, Heidelberg, 2013. Springer Berlin Heidelberg.
\newblock ISBN 978-3-642-40760-4.

\bibitem[McNabb et~al.(2018)McNabb, Tait, McIlwain, Anderson, Suckling, Kydd,
  and Russell]{Carolyn2018Functional}
C.~B. McNabb, R.~J. Tait, M.~E. McIlwain, V.~M. Anderson, J.~Suckling, R.~R.
  Kydd, and B.~R. Russell.
\newblock Functional network dysconnectivity as a biomarker of treatment
  resistance in schizophrenia.
\newblock \emph{Schizophrenia Research}, 195:\penalty0 160--167, 2018.
\newblock ISSN 0920-9964.
\newblock \doi{https://doi.org/10.1016/j.schres.2017.10.015}.
\newblock URL
  \url{https://www.sciencedirect.com/science/article/pii/S0920996417306242}.

\bibitem[Menon(2013)]{Menon2013Developmental}
V.~Menon.
\newblock Developmental pathways to functional brain networks: emerging
  principles.
\newblock \emph{Trends in Cognitive Sciences}, 17\penalty0 (12):\penalty0
  627–640, 2013.
\newblock \doi{10.1016/j.tics.2013.09.015}.

\bibitem[Moradi et~al.(2020)Moradi, Berangi, and Minaei]{moradi2020survey}
R.~Moradi, R.~Berangi, and B.~Minaei.
\newblock A survey of regularization strategies for deep models.
\newblock \emph{Artificial Intelligence Review}, 53\penalty0 (6):\penalty0
  3947--3986, 2020.
\newblock \doi{10.1007/s10462-019-09784-7}.

\bibitem[Naitzat et~al.(2020)Naitzat, Zhitnikov, and Lim]{Naitzat2020Topology}
G.~Naitzat, A.~Zhitnikov, and L.-H. Lim.
\newblock Topology of deep neural networks.
\newblock \emph{Journal of Machine Learning Research}, 21\penalty0
  (184):\penalty0 1--40, 2020.
\newblock URL \url{http://jmlr.org/papers/v21/20-345.html}.

\bibitem[Nayak(2009)]{Richi2009Generating}
R.~Nayak.
\newblock Generating rules with predicates, terms and variables from the pruned
  neural networks.
\newblock \emph{Neural Networks}, 22\penalty0 (4):\penalty0 405--414, 2009.
\newblock ISSN 0893-6080.
\newblock \doi{https://doi.org/10.1016/j.neunet.2009.02.001}.
\newblock URL
  \url{https://www.sciencedirect.com/science/article/pii/S0893608009000161}.

\bibitem[Park and Kwak(2017)]{Park2017Analysis}
S.~Park and N.~Kwak.
\newblock Analysis on the dropout effect in convolutional neural networks.
\newblock In S.-H. Lai, V.~Lepetit, K.~Nishino, and Y.~Sato, editors,
  \emph{Computer Vision -- ACCV 2016}, pages 189--204, Cham, 2017. Springer
  International Publishing.
\newblock ISBN 978-3-319-54184-6.

\bibitem[Petri et~al.(2014)Petri, Expert, Turkheimer, Carhart-Harris, Nutt,
  Hellyer, and Vaccarino]{Petri2014Homological}
G.~Petri, P.~Expert, F.~Turkheimer, R.~Carhart-Harris, D.~Nutt, P.~J. Hellyer,
  and F.~Vaccarino.
\newblock Homological scaffolds of brain functional networks.
\newblock \emph{Journal of The Royal Society Interface}, 11\penalty0
  (101):\penalty0 20140873, 2014.
\newblock \doi{10.1098/rsif.2014.0873}.
\newblock URL
  \url{https://royalsocietypublishing.org/doi/abs/10.1098/rsif.2014.0873}.

\bibitem[Raghu et~al.(2017)Raghu, Poole, Kleinberg, Ganguli, and
  Sohl-Dickstein]{Maithra2017On}
M.~Raghu, B.~Poole, J.~Kleinberg, S.~Ganguli, and J.~Sohl-Dickstein.
\newblock On the expressive power of deep neural networks.
\newblock In D.~Precup and Y.~W. Teh, editors, \emph{Proceedings of the 34th
  International Conference on Machine Learning}, volume~70 of \emph{Proceedings
  of Machine Learning Research}, pages 2847--2854. PMLR, 06--11 Aug 2017.
\newblock URL \url{https://proceedings.mlr.press/v70/raghu17a.html}.

\bibitem[Ribeiro et~al.(2016)Ribeiro, Singh, and Guestrin]{2016why}
M.~T. Ribeiro, S.~Singh, and C.~Guestrin.
\newblock "why should i trust you?": Explaining the predictions of any
  classifier.
\newblock In \emph{the 22nd ACM SIGKDD International Conference}, 2016.

\bibitem[Rieck et~al.(2019)Rieck, Togninalli, Bock, Moor, Horn, Gumbsch, and
  Borgwardt]{Rieck2019Neural}
B.~Rieck, M.~Togninalli, C.~Bock, M.~Moor, M.~Horn, T.~Gumbsch, and
  K.~Borgwardt.
\newblock Neural persistence: A complexity measure for deep neural networks
  using algebraic topology.
\newblock In \emph{7th International Conference on Learning Representations,
  {ICLR} 2019, New Orleans, LA, USA, May 6-9, 2019}. OpenReview.net, 2019.
\newblock URL \url{https://openreview.net/forum?id=ByxkijC5FQ}.

\bibitem[Ringo et~al.(1994)Ringo, Doty, Demeter, and Simard]{Ringo1994Time}
J.~L. Ringo, R.~W. Doty, S.~Demeter, and P.~Y. Simard.
\newblock {Time Is of the Essence: A Conjecture that Hemispheric Specialization
  Arises from Interhemispheric Conduction Delay}.
\newblock \emph{Cerebral Cortex}, 4\penalty0 (4):\penalty0 331--343, 07 1994.
\newblock ISSN 1047-3211.
\newblock \doi{10.1093/cercor/4.4.331}.
\newblock URL \url{https://doi.org/10.1093/cercor/4.4.331}.

\bibitem[Rudie et~al.(2013)Rudie, Brown, Beck-Pancer, Hernandez, Dennis,
  Thompson, Bookheimer, and Dapretto]{Rudie2013Altered}
J.~Rudie, J.~Brown, D.~Beck-Pancer, L.~Hernandez, E.~Dennis, P.~Thompson,
  S.~Bookheimer, and M.~Dapretto.
\newblock Altered functional and structural brain network organization in
  autism.
\newblock \emph{NeuroImage: Clinical}, 2:\penalty0 79--94, 2013.
\newblock ISSN 2213-1582.
\newblock \doi{https://doi.org/10.1016/j.nicl.2012.11.006}.
\newblock URL
  \url{https://www.sciencedirect.com/science/article/pii/S2213158212000356}.

\bibitem[Ryali et~al.(2012)Ryali, Chen, Supekar, and
  Menon]{Srikanth2012Estimation}
S.~Ryali, T.~Chen, K.~Supekar, and V.~Menon.
\newblock Estimation of functional connectivity in fmri data using stability
  selection-based sparse partial correlation with elastic net penalty.
\newblock \emph{NeuroImage}, 59\penalty0 (4):\penalty0 3852--3861, 2012.
\newblock ISSN 1053-8119.
\newblock \doi{https://doi.org/10.1016/j.neuroimage.2011.11.054}.
\newblock URL
  \url{https://www.sciencedirect.com/science/article/pii/S105381191101336X}.

\bibitem[Shnier et~al.(2019)Shnier, Voineagu, and
  Voineagu]{Shnier2019Persistent}
D.~Shnier, M.~A. Voineagu, and I.~Voineagu.
\newblock Persistent homology analysis of brain transcriptome data in autism.
\newblock \emph{Journal of the Royal Society, Interface}, 16\penalty0 (158),
  2019.
\newblock \doi{10.1098/rsif.2019.0531}.

\bibitem[Singh et~al.(2008)Singh, Memoli, Ishkhanov, Sapiro, Carlsson, and
  Ringach]{Singh2008Topological}
G.~Singh, F.~Memoli, T.~Ishkhanov, G.~Sapiro, G.~Carlsson, and D.~L. Ringach.
\newblock {Topological analysis of population activity in visual cortex}.
\newblock \emph{Journal of Vision}, 8\penalty0 (8):\penalty0 11--11, 06 2008.
\newblock ISSN 1534-7362.
\newblock \doi{10.1167/8.8.11}.
\newblock URL \url{https://doi.org/10.1167/8.8.11}.

\bibitem[Srivastava et~al.(2014)Srivastava, Hinton, Krizhevsky, Sutskever, and
  Salakhutdinov]{Srivastava2014Dropout}
N.~Srivastava, G.~Hinton, A.~Krizhevsky, I.~Sutskever, and R.~Salakhutdinov.
\newblock Dropout: A simple way to prevent neural networks from overfitting.
\newblock \emph{Journal of Machine Learning Research}, 15\penalty0
  (1):\penalty0 1929--1958, 2014.

\bibitem[Stam(2004)]{Stam2004Functional}
C.~Stam.
\newblock Functional connectivity patterns of human magnetoencephalographic
  recordings: a ‘small-world’ network?
\newblock \emph{Neuroscience Letters}, 355\penalty0 (1):\penalty0 25--28, 2004.
\newblock ISSN 0304-3940.
\newblock \doi{https://doi.org/10.1016/j.neulet.2003.10.063}.
\newblock URL
  \url{https://www.sciencedirect.com/science/article/pii/S0304394003012722}.

\bibitem[Strogatz(2001)]{Strogatz2001Exploring}
S.~H. Strogatz.
\newblock Exploring complex networks.
\newblock \emph{Nature}, 410\penalty0 (6825):\penalty0 268--276, mar 2001.
\newblock \doi{10.1038/35065725}.

\bibitem[Štrumbelj et~al.(2009)Štrumbelj, Kononenko, and {Robnik
  Šikonja}]{STRUMBELJ2009886}
E.~Štrumbelj, I.~Kononenko, and M.~{Robnik Šikonja}.
\newblock Explaining instance classifications with interactions of subsets of
  feature values.
\newblock \emph{Data \& Knowledge Engineering}, 68\penalty0 (10):\penalty0
  886--904, 2009.
\newblock ISSN 0169-023X.
\newblock \doi{https://doi.org/10.1016/j.datak.2009.01.004}.
\newblock URL
  \url{https://www.sciencedirect.com/science/article/pii/S0169023X09000056}.

\bibitem[Varoquaux et~al.(2013)Varoquaux, Schwartz, Pinel, and
  Thirion]{Varoquaux2013Cohort}
G.~Varoquaux, Y.~Schwartz, P.~Pinel, and B.~Thirion.
\newblock Cohort-level brain mapping: Learning cognitive atoms to single out
  specialized regions.
\newblock In J.~C. Gee, S.~Joshi, K.~M. Pohl, W.~M. Wells, and L.~Z{\"o}llei,
  editors, \emph{Information Processing in Medical Imaging}, pages 438--449,
  Berlin, Heidelberg, 2013. Springer Berlin Heidelberg.
\newblock ISBN 978-3-642-38868-2.

\bibitem[Watanabe and Yamana(2021)]{Watanabe2021Topological}
S.~Watanabe and H.~Yamana.
\newblock Topological measurement of deep neural networks using persistent
  homology.
\newblock \emph{Annals of Mathematics and Artificial Intelligence}, 2021.
\newblock \doi{10.1007/s10472-021-09761-3}.

\bibitem[Xiao et~al.(2017)Xiao, Rasul, and Vollgraf]{Han2017Fashion}
H.~Xiao, K.~Rasul, and R.~Vollgraf.
\newblock Fashion-mnist: a novel image dataset for benchmarking machine
  learning algorithms, 2017.

\bibitem[Yamins et~al.(2014)Yamins, Hong, Cadieu, Solomon, Seibert, and
  DiCarlo]{Yamins2014Performance}
D.~L.~K. Yamins, H.~Hong, C.~F. Cadieu, E.~A. Solomon, D.~Seibert, and J.~J.
  DiCarlo.
\newblock Performance-optimized hierarchical models predict neural responses in
  higher visual cortex.
\newblock \emph{Proceedings of the National Academy of Sciences}, 111\penalty0
  (23):\penalty0 8619--8624, 2014.
\newblock \doi{10.1073/pnas.1403112111}.
\newblock URL \url{https://www.pnas.org/content/early/2014/05/08/1403112111}.

\bibitem[Yang et~al.(2019)Yang, Joglekar, Song, Newsome, and
  Wang]{Yang2019Task}
G.~R. Yang, M.~R. Joglekar, H.~F. Song, W.~T. Newsome, and X.-J. Wang.
\newblock Task representations in neural networks trained to perform many
  cognitive tasks.
\newblock \emph{Nature Neuroscience}, 22\penalty0 (2):\penalty0 297–306,
  2019.
\newblock ISSN 1546-1726.
\newblock \doi{10.1038/s41593-018-0310-2}.

\bibitem[Young et~al.(2000)Young, Stephan, Hilgetag, Burns, O'Neill, Young, and
  Kotter]{Young2000Computational}
M.~P. Young, K.~E. Stephan, C.~Hilgetag, G.~A. P.~C. Burns, M.~A. O'Neill,
  M.~P. Young, and R.~Kotter.
\newblock Computational analysis of functional connectivity between areas of
  primate cerebral cortex.
\newblock \emph{Philosophical Transactions of the Royal Society of London.
  Series B: Biological Sciences}, 355\penalty0 (1393):\penalty0 111--126, 2000.
\newblock \doi{10.1098/rstb.2000.0552}.
\newblock URL
  \url{https://royalsocietypublishing.org/doi/abs/10.1098/rstb.2000.0552}.

\bibitem[Zeiler and Fergus(2014)]{Zeiler2014Visualizing}
M.~D. Zeiler and R.~Fergus.
\newblock Visualizing and understanding convolutional networks.
\newblock In D.~Fleet, T.~Pajdla, B.~Schiele, and T.~Tuytelaars, editors,
  \emph{Computer Vision -- ECCV 2014}, pages 818--833, Cham, 2014. Springer
  International Publishing.
\newblock ISBN 978-3-319-10590-1.

\bibitem[Zhang et~al.(2021)Zhang, Bengio, Hardt, Recht, and
  Vinyals]{Zhang2021Understanding}
C.~Zhang, S.~Bengio, M.~Hardt, B.~Recht, and O.~Vinyals.
\newblock Understanding deep learning (still) requires rethinking
  generalization.
\newblock \emph{Commun. ACM}, 64\penalty0 (3):\penalty0 107–115, Feb. 2021.
\newblock ISSN 0001-0782.
\newblock \doi{10.1145/3446776}.
\newblock URL \url{https://doi.org/10.1145/3446776}.

\end{thebibliography}

\end{document}